\documentclass{article}

    \PassOptionsToPackage{numbers, compress}{natbib}

\usepackage[hidelinks,colorlinks]{hyperref}
\usepackage[preprint]{neurips_2025}

\usepackage[svgnames,table]{xcolor}

\usepackage{multirow}
\usepackage[utf8]{inputenc}
\usepackage[T1]{fontenc}
\usepackage{microtype}
\usepackage{graphicx}
\usepackage{subcaption}
\usepackage{amsmath,amssymb}    
\usepackage{booktabs}
\usepackage{nicefrac}
\usepackage{listings}
\usepackage{url}
\usepackage{float}
\usepackage[capitalise]{cleveref}
\usepackage{enumitem}
\usepackage{multirow}
\newcommand{\tba}[1][]{\textcolor{Blue}{TBA\ifx&#1&\else---#1\fi}}
\newcommand{\yes}{\textcolor{DarkGreen}{\checkmark}}
\newcommand{\no}{\textcolor{DarkRed}{$\times$}}

\definecolor{myblue}{rgb}{0,0.3,0.6}
\hypersetup{linkcolor=myblue,urlcolor=myblue,citecolor=myblue,anchorcolor=myblue}
\crefformat{footnote}{#2\footnotemark[#1]#3}
\crefrangeformat{footnote}{#3\footnotemark[#1]#4--#5\footnotemark[#2]#6}
\crefmultiformat{footnote}{#2\footnotemark[#1]#3}{\textsuperscript{,}#2\footnotemark[#1]#3}{\textsuperscript{,}#2\footnotemark[#1]#3}{\textsuperscript{,}#2\footnotemark[#1]#3}
\setlist{leftmargin=2em,topsep=0pt,partopsep=0pt,parsep=0pt,itemsep=3pt}
\usepackage{enumitem}
\setlist[itemize]{noitemsep, topsep=0pt, leftmargin=1.5em}
\usepackage{float}
\newcommand{\tabfn}[1][]{{\ifx&#1&\textsuperscript{$*$}\else\textsuperscript{\myfnsymbol{#1}}\fi}}
\newcommand{\myfnsymbol}[1]{\ensuremath{\ifcase#1 \or * \or \dagger \or \ddagger \or \mathsection \or \mathparagraph \else \@ctrerr \fi}}

\title{REGen: Multimodal Retrieval-Embedded Generation for Long-to-Short Video Editing}

\author{
  Weihan Xu\textsuperscript{1}
  \quad Yimeng Ma\textsuperscript{1}
  \quad Jingyue Huang\textsuperscript{2}
  \quad Yang Li\textsuperscript{1}
  \quad Wenye Ma\textsuperscript{3}
  \\
  \textbf{Taylor Berg-Kirkpatrick\textsuperscript{2}}
  \quad \textbf{Julian McAuley\textsuperscript{2}}
  \quad \textbf{Paul Pu Liang\textsuperscript{4}}
  \quad \textbf{Hao-Wen Dong\textsuperscript{5}}
  \\
  \textsuperscript{1}Duke University
  \quad
  \textsuperscript{2}University of California, San Diego
  \quad
  \textsuperscript{3}MBZUAI
  \quad\\
  \textsuperscript{4}MIT
  \quad
  \textsuperscript{5}University of Michigan
  \\
  \texttt{\{weihan.xu,yimeng.ma,yang.li\}@duke.edu},
  \;\\
  \texttt{\{jih150,tberg,jmcauley\}@ucsd.edu},
  \;\\
  \texttt{wenye.ma@mbzuai.ac.ae},
  \;\\
  \texttt{ppliang@mit.edu},
  \;
  \texttt{hwdong@umich.edu}
}

\begin{document}

\maketitle

\begin{abstract}
Short videos are an effective tool for promoting contents and improving knowledge accessibility. While existing extractive video summarization methods struggle to produce a coherent narrative, existing abstractive methods cannot `quote' from the input videos, i.e., inserting short video clips in their outputs. In this work, we explore novel video editing models for generating shorts that feature a coherent narrative with embedded video insertions extracted from a long input video. We propose a novel retrieval-embedded generation framework that allows a large language model to quote multimodal resources while maintaining a coherent narrative. Our proposed \textit{REGen} system first generates the output story script with quote placeholders using a finetuned large language model, and then uses a novel retrieval model to replace the quote placeholders by selecting a video clip that best supports the narrative from a pool of candidate quotable video clips. We examine the proposed method on the task of documentary teaser generation, where short interview insertions are commonly used to support the narrative of a documentary. Our objective evaluations show that the proposed method can effectively insert short video clips while maintaining a coherent narrative. In a subjective survey, we show that our proposed method outperforms existing abstractive and extractive approaches in terms of coherence, alignment, and realism in teaser generation.
\end{abstract}


\section{Introduction}
Generating shorts from long videos allows audiences to digest information in a more engaging way and helps content creators promote their original contents.
Unlike text or visual-only summaries, short videos with visuals and audio are more engaging \cite{zhang2024unveilingimpactmultimodalinteractions}, accelerate comprehension \cite{comprehension_enhance}, and improve recommendation and search \cite{liu2024recgpt4vmultimodalrecommendationlarge}. Existing approaches for producing shorts from long videos can be categorized into extractive or abstractive methods. Extractive methods stitch together video clips extracted from the input video, yet this may produce disjointed videos that do not together convey a coherent story \cite{he2023alignattendmultimodalsummarization,textclue1,textclue2,textclue3,audiovisualclue1,audiovisualclue2,audiovisualclue3}. In contrast, abstractive approaches synthesize new narratives \cite{xu2024teasergengeneratingteaserslong} and even new scenes \cite{dalal2025oneminutevideogenerationtesttime}, but these methods cannot insert extracted video clips from the input video to support the generated narrative.
Moreover, while recent retrieval-augmented generation (RAG) methods can augment a large language model (LLM) with additional knowledge at inference time, these methods fail to quote multimodal materials from external sources and embed the exact quotes into their outputs, and they sometimes fabricate or misattribute content when faced with extended contexts \cite{llm_hall,mllm_hall,xu2025hallucinationinevitableinnatelimitation}.

 Crafting effective short videos requires both creating a coherent narrative and grounding it with raw material extracts,
especially for domains that necessitate strong factualness and reliability such as journalism and education. Further, there exists no video dataset with annotation that identifies externally quoted footage from original narrative segments, making it hard to approach this task through a data-driven approach.

\begin{figure}
  \centering
  \includegraphics[width=\textwidth]{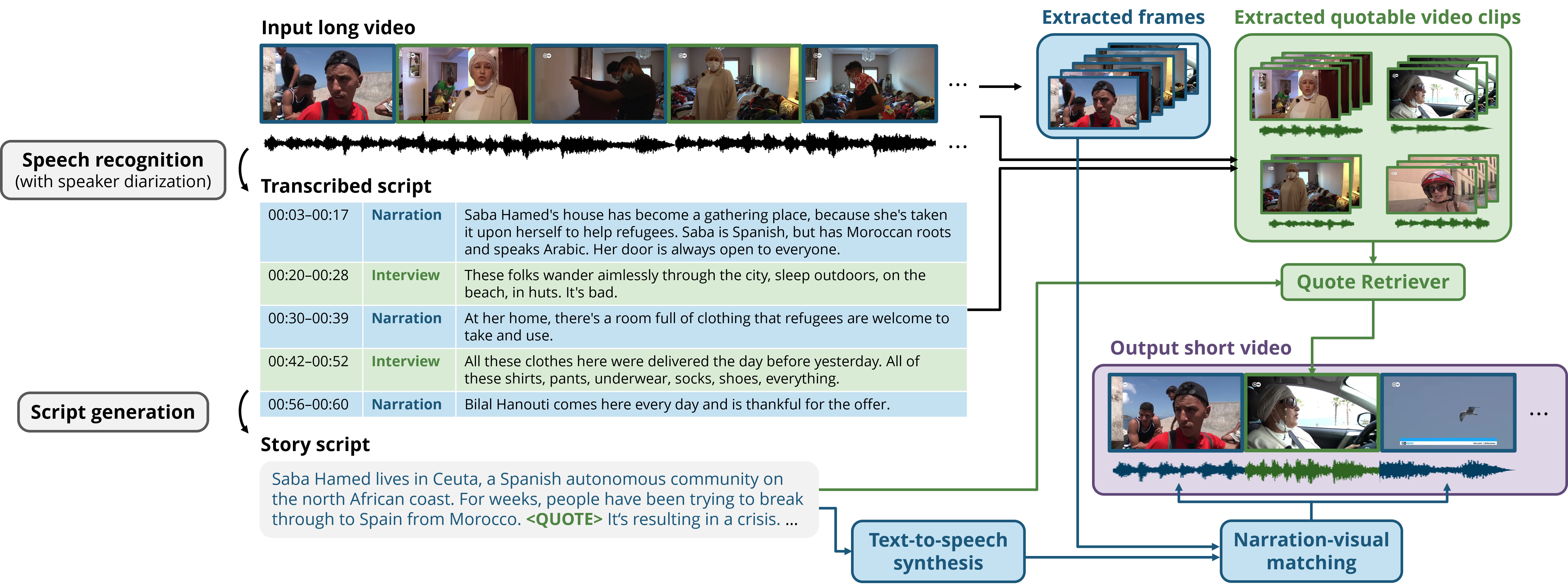}
  \caption{An overview of the proposed \textit{REGen} system for long-to-short video editing. Given a long input video, we first transcribe the narrations and dialogues using a pretrained automatic speech recognition model, and then we use a finetuned large language model to generate the output story script with quote placeholders (i.e., the \texttt{<QUOTE>} token). For the generated narration, following \cite{xu2024teasergengeneratingteaserslong}, we first synthesize the narration into audio using a text-to-speech synthesis model and apply a narration-visual matching algorithm to find accompanying visuals. For the generated quote placeholders, we propose an encoder-decoder based \textit{Quote Retriever} to select a video clip that best supports the narrative from a pool of quotable video clips extracted from the input video. The proposed system represents a new hybrid video editing model that combines abstractive and extractive methods.}
  \label{fig:overview}
\end{figure}

In this work, we introduce \textit{REGen}, a novel multimodal retrieval-embedded generation framework for editing long videos into shorts (see \cref{fig:overview}). 
In the first script generation stage, we finetune an LLM to generate story scripts with quote placeholders that will be fulfilled later in the second stage.
In the second quotation retrieval stage, we then train a multitask encoder-decoder language model to select a video clip extracted from the input video so that it can support the narrative. For other generated narration, we follow \cite{xu2024teasergengeneratingteaserslong} to synthesize the narration and accompanying it with visuals selected from frames extracted from the input video.
To train the proposed system, we use the DocumentaryNet dataset \cite{xu2024teasergengeneratingteaserslong} and construct training samples with transcribed, timestamped narrations and quotable interviews using an existing speech transcription and speaker diarization model \cite{whisperx}.

We conduct extensive experiments to evaluate the effectiveness of our proposed models through objective evaluation metrics and a subjective survey. We show that the proposed REGen models can effectively edit a long documentary into a short teaser that has a coherent narrative and contain video quotations that support the narrative. Our experimental results show that our proposed method outperforms several abstractive and extractive baseline models in terms of coherence, audiovisual alignment, and realism.
Video samples and all source code can be found on our website.\footnote{\label{fn:demo}\url{https://wx83.github.io/REGen/}}



Our contribution can be summarized as follows:

\begin{itemize}
  \item We propose a new retrieval-embedded generation framework that allows an LLM to quote multimodal resources while maintaining a coherent narrative.
  \item We propose a novel long-to-short video editing model for generating shorts that feature a coherent narrative with embedded video insertions extracted from a long input video.
\end{itemize}


\section{Related Work}

\paragraph{Generative modeling with factual grounding}

Previous work on generative modeling with factual grounding is primarily in the text domain and falls into two main categories: attribution-aware LLMs \cite{attributellm1,finegrainattribute} and retrieval-augmented generation (RAG) methods \cite{rag1,rag2,rag3}. Attribution-aware LLMs enhance verifiability by generating responses with in-text citations or via post-hoc attributions. In addition, they employ coarse attributions such as URLs \cite{thoppilan2022lamdalanguagemodelsdialog} or document identifiers\cite{liu-etal-2023-evaluating}. RAG-based approaches first retrieve relevant documents or text chunks and then condition the generation on the retrieved passages \cite{rag1,rag2,rag3}. In contrast, our work targets multimodal outputs and performs exact quoting from external multimodal resources, generating narratives that directly quote raw quotable footage as grounding evidence to support the generated narratives.


\begin{table}
    \footnotesize
    \centering
    \caption{Comparison of related methods in trailer generation and multimodal summarization}
    \vspace{.5ex}
    \label{tab:method}
\begin{tabular}{lcccccccccc}
  \toprule
   & \multicolumn{2}{c}{Type}
   & \multicolumn{3}{c}{Input modality}
   & \multicolumn{3}{c}{Output modality} &\multirow{2}{*}[-1ex]{\shortstack{Video clip\\insertions\tabfn[1]}}\\
  \cmidrule(lr){2-3} \cmidrule(lr){4-6} \cmidrule(lr){7-9}
  Model 
         & Ext.\phantom{.} 
         & Abs. 
         & Frames 
         & Video 
         & Narr. 
         & Text 
         & Video 
         & Frames \\ 
        \midrule
        CLIP-IT \cite{NEURIPS2021_7503cfac} &\yes & \no & \yes &\yes  &\no  &\no &\yes & \yes & \no\\
        A2Summ \cite{he2023alignattendmultimodalsummarization} &\yes & \no & \yes &\yes  &\yes  & \yes & \yes & \yes  & \yes\\
        LfVS \cite{argaw2024scalingvideosummarizationpretraining} &\yes & \no& \yes & \yes  & \yes  & \yes & \yes & \yes  & \yes \\
        TGT \cite{argaw2024automatedmovietrailergeneration} &\yes & \no& \yes & \yes  & \no & \no & \yes & \yes  & \no\\
        TaleSumm \cite{singh2024previouslyrecapsstory} &\yes & \no &\yes & \yes  & \yes  &\yes &\yes & \yes & \yes \\
        VTSUM-BLIP \cite{lin2023videoxum} &\no & \yes &\yes &\yes & \no  &\yes\tabfn[2] & \yes &\yes  & \no\\
        TeaserGen \cite{xu2024teasergengeneratingteaserslong}  &\no & \yes &\yes & \yes & \yes  &\yes &\yes & \yes & \no \\
        
    \midrule
        REGen \cite{xu2024teasergengeneratingteaserslong}  &\yes & \yes &\yes & \yes  & \yes  &\yes &\yes & \yes  & \yes\\
        \bottomrule
    \end{tabular}\\[.5ex]
    \tabfn[1] Whether its outputs include extracted video clips with original sounds\quad
    \tabfn[2] Achieved by dense video captioning
    \vspace{-1ex}
\end{table}

\paragraph{Long-to-short Video Editing}

Long-to-short video editing like video summarization or trailer generation addresses the challenge of condensing long‐form videos into informative short videos. Prior work can be broadly divided into extractive and abstractive approaches. Extractive methods identify and splice together key clips directly from the source footage. For example, A2Summ \citep{he2023alignattendmultimodalsummarization} produces extractive summaries with a unified multimodal transformer‐based model to predict key sentences and their time‐aligned video segments. LfVS \cite{argaw2024scalingvideosummarizationpretraining} utilizes large language models (LLMs) to extract key sentences from transcribed text, which are then paired with time‐aligned video segments to create pseudo‐ground‐truth summaries. TaleSumm \cite{singh2024previouslyrecapsstory} introduces a two‐level hierarchical model that identifies important sub‐stories in a TV episode narrative. Although these techniques preserve the authenticity of the original clips, they often yield a disjointed viewing experience due to abrupt transitions between extracted segments.
Abstractive methods, by contrast, first generate a cohesive narrative script and then retrieve or synthesize matching visuals. For instance, TeaserGen \cite{xu2024teasergengeneratingteaserslong} prompts a large language model to produce a teaser script and subsequently fetch corresponding video clips. VTSUM-BLIP \cite{lin2023videoxum} jointly train parallel video and text summarization decoders, enabling end‐to‐end video‐to‐video summarization. Though abstractive approaches deliver smoother, more story‐like short videos, they risk drifting from factual grounding. In this work, we propose a hybrid framework that automatically generates a coherent narrative and seamlessly inserts extractive quotable segments as grounding evidence. We compare related methods in \cref{tab:method}.

\section{Method}

To generate short videos that quote contents from long videos, we adopt a two-stage method: first, we generate scripts with explicit quotation encoding; then we retrieve the corresponding quotable segments from long videos to fulfill each quotation coherently and support the surrounding narration.
\subsection{Generating Script with Quotation via Fine‐Tuned LLaMA}

To train an LLM to identify quote insertion points, we leverage ASR with speaker diarization (see \cref{sec:script_gen}) to generate data with quotable segments separated from narration. We then explore two quote encodings for finetuning a pretrained language model to enable quoting from long contexts:
\begin{align}\label{sec:quote_encoding}
&\mathrm{REGen\text{-}DQ\ (direct\ quote)}: &&\dots, x_i, \texttt{<SOQ>}, y_1, \dots, y_n, \texttt{<EOQ>}, x_{i+1}, \dots\\
&\mathrm{REGen\text{-}IDQ\ (indirect\ quote)}: &&\dots, x_i, \texttt{<QUOTE>}, x_{i+1}, \dots
\end{align}
To reduce hallucinations and ensure comprehensive coverage of the source material, we transcribe the documentary audio with WhisperX \cite{whisperx}, split it into ten chunks, and use GPT-4o \cite{openai2024gpt4technicalreport} to generate one-sentence summary for each chunk as inputs to LLM. Specially, we finetune LLaMA \cite{touvron2023llamaopenefficientfoundation} using instruction finetuning. Our template can be found in \cref{sec:llama_finetune_regen_dq,sec:llama_finetune_regen_idq}.
 
For the generated narrations, following TeaserGen \cite{xu2024teasergengeneratingteaserslong}, we accompany the synthesized narration with visuals using an interval-based matching approach, extracting content corresponding to each narration sentence. We use the pretrained UniVTG model \cite{lin2023univtg} to identify highlights aligned with the narrations.
For the quotable part, we search our quotable clip base for the sentence embedding closest to the text wrapped in \texttt{<SOQ>} and \texttt{<EOQ>} for direct quotes. We will refer to this model as \textbf{REGen-DQ}. For indirect quotes, we propose a retriever module to retrieve quotable segments that blend with the surrounding narration (see \cref{sec:retriever_module}), which we will refer to as \textbf{REGen-IDQ}.

\begin{figure}
  \centering
  \includegraphics[width=.9\textwidth]{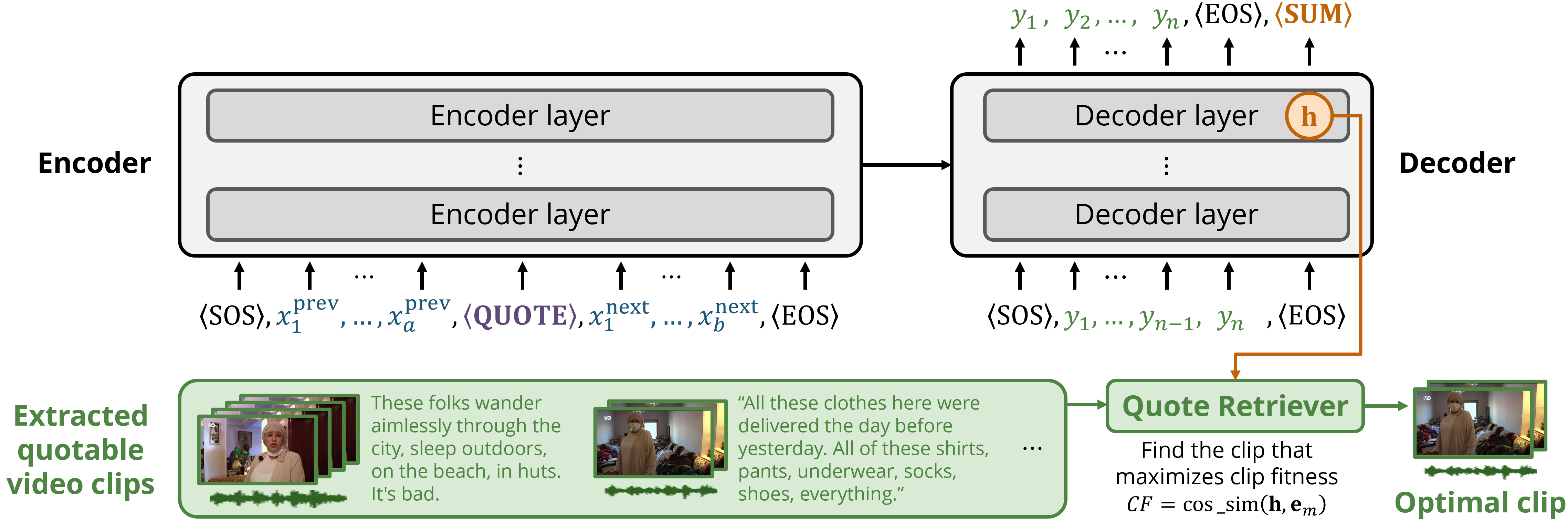}
  \caption{An illustration of the proposed two-stage quote retriever REGen-IDQ. We finetune an encoder-decoder language model that learns to 1) fulfill quotation placeholders (i.e., the \texttt{<QUOTE>} token) and 2) produce an embedding vector $\mathbf{h}$ that summarizes the quotation content. We then use the embedding vector $\mathbf{h}$ as the query to retrieve a video clip from a pool of candidate quotable video clips extracted from the input video. The optimal clip is selected based on our proposed \textit{clip fitness} measure (see \cref{sec:retriever_module} for its definition). In this work, we consider all non-narrator video clips as candidates for the quote retriever. Note that this framework can be generalized to support quoting materials in any modality such as audio and images as long as we can find a proper fitness measure.}
  \label{fig:retriever}
\end{figure}

\subsection{Retrieving Quotable Segments Aligned with Narrative Context for RGEen-IDQ}
\label{sec:retriever_module}

As shown in \cref{fig:retriever}, to retrieve a quotable segment from a curated database that supports the surrounding narrative, we frame this task as a multitask learning problem: the proposed quote retriever is trained jointly with a masked language modeling loss and a retrieval loss.

To fulfill the content of the quotation placeholders generated by REGen-IDQ,
we finetune a pretrained encoder-decoder model BART \cite{lewis-etal-2020-bart} to perform masked infilling conditioned on surrounding narrations. 
Let previous sentence be $s^\mathrm{prev} = (x^\mathrm{prev}_1, \dots, x^\mathrm{prev}_a)$ and the following sentence be $s^{next} = (x^\mathrm{next}_1, \dots, x^\mathrm{next}_b)$. We have our encoder input and decoder output as
\begin{align}
&\mathrm{Input}: &&\texttt{<SOS>}, x^\mathrm{prev}_1, \dots, x^\mathrm{prev}_a, \texttt{<QUOTE>}, x^\mathrm{next}_1, \dots, x^\mathrm{next}_b, \texttt{<EOS>}\\
&\mathrm{Output}: &&\texttt{<SOS>}, y_1, \ldots, y_n, \texttt{<EOS>}, \texttt{<SUM>}
\end{align}
The proposed method decodes meaningful sequence that supporting nearby narrations, while the added special \texttt{<SUM>} token is expected to summarize the content of the quatation.



Inspired by retrieval-augmented generation (RAG) \cite{rag1,rag2,rag3}, we propose a retrieval module to find a suitable quotable segment from a candidate pool that best matches the decoded sentences. We use the hidden state of the final decoder token (i.e., the special \texttt{<SUM>} token) from each generated sequence to retrieve quotable video segments from the candidate pool. For each \texttt{<QUOTE>} placeholder and a pool of candidate quotable video clips $C = \{c_1, \dots, c_M\}$, we retrieve the optimal video clip $c^*$ that maximizes the \textit{clip fitness}, defined as $CF = \mathrm{cos\_sim}(\mathbf{h}, \mathbf{e}_m)$, where $h$ is the last-layer hidden state at the final decoder layer for the last token (i.e., the \texttt{<SUM>} token) and $\mathbf{e}_m$ is an embedding vector that captures the semantic meaning of a candidate clip $c_m$. We consider two variants of $\mathbf{e}_m$: First, we consider $\mathbf{e}_m = f(\mathrm{concat}(\mathbf{e}^\mathrm{text}, \mathbf{e}^\mathrm{img}_m))$, where $f$ is a learnable mapping parameterized as a two-layer multi-layer perceptrons and $\mathbf{e}^\mathrm{text}_1$ is the sentence embedding of the whole narration of the quote video segments. To aggregate visual information, we define $\mathbf{e}^\mathrm{img}_m$ as the concatenated frame embeddings of three randomly selected frames for the quotable segments. The proposed multimodal fusion module $f$ is expected to learn to combine visual and textual information efficiently to optimize the retrieval performance. 
We will refer to this retriever as \textbf{QuoteRetriever-TV}.  
The REGen-IDQ model using this retriever will be referred to as \textbf{REGen-IDQ-TV}. In addition, we consider $\mathbf{e}_m = \mathbf{e}^\mathrm{text}_m$, i.e., a retrieval model that considers only textual information. We will refer to this retriever as \textbf{QuoteRetriever-T} and the corresponding full system as \textbf{REGen-IDQ-T}.



During training,
we jointly train the fusion module and fine-tune the pretrained BART\cite{lewis-etal-2020-bart} model with a multitask loss function: $\label{eq:loss_func}L = L_\mathrm{gen} + \alpha L_\mathrm{ret}$,
where $L_{\mathrm{gen}}$ is the token level cross‐entropy loss for masked language modeling, and $L_\mathrm{ret}$ is the retrieval loss defined as
\begin{equation}
L_{\mathrm{ret}} = -\sum_{k=1}^K \log \frac{\exp\bigl(\mathrm{cos\_sim}(\mathbf{h}_k, \mathbf{e}^*)\bigr)}{\sum_{c_j\in C^-_k}\exp\bigl(\mathrm{cos\_sim}(\mathbf{h}_k, \mathbf{e}_j)\bigr)}\,,
\end{equation}
where $e^*$ is the sentence embedding of the ground truth narration and $C^-_k$ the set of negatives samples for quotation placeholder $k$.
Furthermore, we train the retriever with in‐batch negative sampling, and explore a GroupSampler module to construct hard negative samples, further separating ground-truth quotable video clips from the remaining quotable video clips in the video clip base (see \cref{sec:appendix_groupsampler} for more details).

\section{Experimental Setup}

\subsection{Dataset and Implementation Details}

In this work, we use the DocumentaryNet \cite{xu2024teasergengeneratingteaserslong} dataset in our experiments. DocumentaryNet contains 1,269
documentaries paired with their teasers from three reliable sources: DW Documentary, Public Broadcasting Service (PBS) and National Geographic. We perform speaker diarization to generate scripts using WhisperX \cite{whisperx} on DocumentaryNet and automatically detect narrator segments by assuming that narrations usually correspond to the longest transcribed audio segments. In this work, we consider all non-narrator video clips as quotable video clip candidates for the quote retriever. To validate our data annotations, we recruited four people to validate the ASR and narrator‐identification results. Dataset details can be found in \cref{sec:dataset_processing}. We use an FPS (frame per second) of 1. We also include implementation details in \cref{sec:implementation_detail}. 

\subsection{Script Generation}\label{sec:script_gen_experiment}

In our first experiment, we evaluate the performance of our proposed method in generating coherent scripts with quotations, as well as its ability to conduct exact quotations from input videos.

\paragraph{Baselines} 
We compare our model with script existing multimodal video summarization model \cite{he2023alignattendmultimodalsummarization}, teaser generation model \cite{xu2024teasergengeneratingteaserslong}, and three LLM-based method. 
\begin{itemize}
    \item \textbf{Random Extraction}: We randomly sample sentences from main documentary transcription. 
    \item \textbf{Extractive-then-Smoothing (ETS)}: We select two interviews whose content is closest to the video title and ask GPT-4o to connect the extracted interviews into a cohesive story. We include the prompts in \cref{sec:extractive_then_smoothing}.
    \item \textbf{A2Summ} \cite{he2023alignattendmultimodalsummarization}: This baseline model uses an extractive method to select joint textual and visual segments with temporal correspondence. Since A2Summ can only process videos shorter than 300 seconds, we divide each video into ten chunks, select key segments from each chunk separately, and concatenate them together.
    \item \textbf{TeaserGen} \cite{xu2024teasergengeneratingteaserslong}: This baseline model divides the audio transcription of long videos into ten chunks, requests a one-sentence summary for each, and instructs GPT-4o \cite{openai2024gpt4technicalreport} to weave these summaries into a story-like teaser ending with a compelling question.
    \item \textbf{GPT-4o-DQ}: We prompt GPT-4o \cite{openai2024gpt4technicalreport} to generate an introduction script given a summary of the main documentary content and include in-text quotations. We include the prompt in \cref{sec:gpt_4o_dq_prompt}. 
    \item \textbf{GPT-4o-SP-DQ}: We split the script processed in \cref{sec:script_gen} into ten chunks, and ask GPT-4o\cite{openai2024gpt4technicalreport} to generate a concise, engaging summary for each chunk. We then concatenate these ten summaries to reconstruct the complete script with speaker labeled. We include our prompts in \cref{sec:gpt_4o_idq}.
    \item \textbf{LLaMA-DQ}: We prompt LLaMA\cite{touvron2023llamaopenefficientfoundation} to generate an introduction script given a summary of the main documentary content and include in-text quotations. We include the prompt in \cref{sec:llama_dq_prompt}.
\end{itemize}

\paragraph{Objective Evaluation Metrics}
We evaluate our approach using three primary metrics. 
First, we measure \textit{quotation density index} (\textbf{QDI}), i.e., the average number of quotes inserted per documentary. Second, we compute \textit{quote coverage rate} (\textbf{QCR}), the proportion of test videos in which at least one quotation is correctly inserted. Third, we define \textit{overlap ratio} (\textbf{OR}) to measure the overlap between direct quoted contents by large language model with the ground truth interviews as $\mathrm{OR} = \#\{\text{overlap words}\} / \#\{\text{words in matched interviews}\}$.

\subsection{Quote Retriever Evaluation}
In this experiment, we evaluate whether the proposed retriever can retrieve the correct video clip in our test set and assess its generalizability when applied to LLM generated scripts.

\paragraph{Baselines}
To evaluate the effectiveness of our proposed quote retriever, we compare it against two baselines: random selection and GPT-based infilling of \texttt{<QUOTE>} with surrounding narrations.
\begin{itemize}
    \item \textbf{Random Selection}: Randomly choose interview segments for insertion.
    \item \textbf{GPT-4o Infilling}: Given the preceding and succeeding narration chunks, we prompt GPT-4o \cite{openai2024gpt4technicalreport} to generate content to fill the \texttt{<QUOTE>} position, and then retrieve the nearest neighbor in the sentence embedding space \cite{sentence-transformers-all-mpnet-base-v2} from our interview base.
\end{itemize}
\paragraph{Evaluation Metrics} \label{sec:subjective_survey}
For objective evaluation, we evaluate our retrieval stage using recall, reporting Recall@1, Recall@5 and Recall@10 on teasers in the test set. Recall@5 indicates that the correct segment appears among the top five retrieved interviews. 
To further assess the generalizability of our retriever when applied to LLM-generated scripts, we conduct a subjective evaluation. Twenty-one participants (11 evaluating version A and 10 evaluating version B) rate each inserted interview segment in the generated teasers on a five-point Likert scale, judging the effectiveness of the insertion based on how well it supports the surrounding claim and maintains natural flow. We include survey questions in \cref{survey_questions}.

\subsection{Documentary Teaser Generation}\label{sec:documentary_eval}

In this experiment, we measure our model performance in documentary teaser generation task. 

\paragraph{Baselines}
We consider Random Extraction, Extractive-then-Smoothing, A2Summ \cite{he2023alignattendmultimodalsummarization}, and TeaserGen \cite{xu2024teasergengeneratingteaserslong} described in \cref{sec:script_gen_experiment} as baselines. Additionally, for GPT-4o-DQ, we extract quoted segments within quotation marks and use nearest-neighbor retrieval to fetch matching visual clips from the interview pool; for GPT-4o-SP-DQ, we retrieve interview segments from the interview pool via nearest-neighbor search on script segments labeled as non-narrator content.

\paragraph{Evaluation Metrics}

We first measure the quality of the final script, where each \texttt{<QUOTE>} marker generated by the script-generation stage has been replaced by its retrieved interview segment in the retrieval stage. Specifically, we report ROUGE F1 \cite{lin-2004-rouge}, which measures n-gram overlap and sequence continuity between generated and reference teaser scripts. In addition, we assess narrative coherence using G-Eval \cite{liu2023gevalnlgevaluationusing} on the DeepEval platform \cite{deepeval2025}. 
In addition, following TeaserGen~\cite{xu2024teasergengeneratingteaserslong}, we report five retrieval-based metrics: F1, Scene Change Rate (SCR), Repetitiveness (REP), CLIPScore \cite{hessel2022clipscorereferencefreeevaluationmetric}, and VTGHLS \cite{liu2023gevalnlgevaluationusing, xu2024teasergengeneratingteaserslong}. F1 measures retrieval accuracy against the ground-truth teaser, while SCR (the frequency of scene transitions) and REP (the degree of repetitive content) capture aspects of temporal continuity that affect the viewer experience. We report CLIPScore (CLIPS-I and CLIP-N) to measure audiovisual alignment, and VTGHLS to measure the likelihood that each selected frame will be perceived as a highlight relevant to the video title (see \cref{sec:objective_eval_detail} for more details). Moreover, we define the interview ratio as the fraction of interview time (in seconds) in a video.

Following the subjective study in \cref{sec:subjective_survey}, we randomly select ten documentaries from the test set, divide them into two groups of five, and ask participants to evaluate the generated teasers on coherence, alignment, realism, and interview effectiveness using a five-point Likert scale.

\subsection{Ablation Study}

We include our ablation study in \cref{sec:ablation_study}. We evaluate the effects of the the max length of BART tokenizer \cite{lewis-etal-2020-bart} (Section \ref{sec:ablation_context_window}), the alpha parameter for balancing losses (Section \ref{sec:ablation_alpha_loss}), the use of GroupSampler during retriever training (Section \ref{sec:ablation_groupsampler}), the choice of loss function (Section \ref{sec:ablation_loss_func}), and the position of the retrieval token \texttt{<SUM>}. 

\section{Results}\label{sec:results}

\subsection{Script Generation}

In this experiment, we compare our model performance in generating scripts with quotations against the following baselines: GPT-4o-DQ, GPT-4o-SP-DQ, random extraction, A2Summ \cite{he2023alignattendmultimodalsummarization}, and TeaserGen \cite{xu2024teasergengeneratingteaserslong}. First, we evaluate whether LLMs such as LLaMA \cite{touvron2023llamaopenefficientfoundation} and GPT-4o \cite{openai2024gpt4technicalreport} can directly quote from long contexts, and we report our results in \cref{sec:direct_quote_table}. We find that GPT-4o cannot quote exact content from long inputs when fed with the full transcript. In addition, while vanilla LLaMA cannot produce meaningful quotes using quotation marks, the proposed finetuning method with \texttt{<SOQ>} and \texttt{<EOQ>} increases the overlap ratio from 0 to 0.07.
Second, we compare our model to GPT-4o-DQ, GPT-4o-SP-DQ, Random Extraction, A2Summ \cite{he2023alignattendmultimodalsummarization}, and TeaserGen \cite{xu2024teasergengeneratingteaserslong} using the quotation density index (QDI) and quote coverage rate (QCR). In \cref{tab:objective_screenplay}, we find that REGen-DQ achieves QCR and QDI values closest to those of the ground truth, indicating that REGen-DQ generates scripts with a similar quote distribution to the ground truth scripts.

\begin{table}
  \caption{Objective evaluation results for documentary teaser script generation}
  \label{tab:objective_screenplay}
  \footnotesize
  \centering
  \vspace{.5ex}
  \begin{tabular}{lcccccccc}
  \toprule
    & \multicolumn{3}{c}{Before fulfillment} 
      & \multicolumn{5}{c}{After fulfillment} \\
    \cmidrule(lr){2-4} \cmidrule(lr){5-9} 
    Model  &Tokens & QCR (\%) & QDI &Tokens & R-1 & R-2 & R-L & G-Eval \\
    \midrule
    Random extraction & - &  98 &  11.71 & 235  & 0.27 & 0.04  &  0.12  & 0.56 $\pm$ 0.02 \\
    ETS  & - & 96 & 1.96  & 340 & 0.21 & 0.03 & 0.11 & 0.81 $\pm$ 0.01 \\
    A2Summ \cite{he2023alignattendmultimodalsummarization}& - & 96 & 3.98  & 172 & 0.27 & 0.03 & 0.13 & 0.42 $\pm$ 0.01 \\  
    TeaserGen \cite{xu2024teasergengeneratingteaserslong}   & -   &- &- & 304 & 0.21 & 0.03 & 0.11 & 0.85 $\pm$ 0.01\\ 
    GPT-4o-DQ & 292 & 98 & 4.02 & 402&  0.22 &0.05 & 0.12 & 0.77 $\pm$ 0.01\\
    GPT-4o-SP-DQ & 631 & 100 & 22.33  & 1372 & 0.13 & 0.03 & 0.07  & 0.75 $\pm$ 0.01\\
    \midrule  
    REGen-DQ & 153  & \textbf{76} & \textbf{2.31} & 210& \textbf{0.28} & \textbf{0.05}& \textbf{0.13}  & 0.43 $\pm$ 0.02 \\
     REGen-IDQ-T  & 98  & 67 & 1.98 & 172  & 0.25 & 0.04 & 0.13  &0.57 $\pm$ 0.02\\
     REGen-IDQ-TV & 98  & 67& 1.98 & 179 & 0.25 &0.04 & 0.13 & \textbf{0.59 $\pm$ 0.01}\\
    \midrule
    Ground truth          &-  & 82 & 3.02 & 121& -   & -   & - &0.62 $\pm$ 0.03\\
    \bottomrule
  \end{tabular}
\end{table}

\begin{table}
  \caption{Comparisons of quote retrieval methods}
  \label{retriver_gold}
  \footnotesize
  \centering
  \vspace{.5ex}
  \begin{tabular}{lccccc}
    \toprule
    Retriever &\shortstack{Similarity\\measure} &\shortstack{Recall@1\\(\%)} & \shortstack{Recall@5\\(\%)} & \shortstack{Recall@10\\(\%)} &\shortstack{Insertion\\effectiveness}\\
    \midrule
    Random & - & 0.00 $\pm$ 0.00 & 0.28 $\pm$ 0.48 &7.22 $\pm$ 5.54  & 3.08 $\pm$ 0.25\\
    GPT-4o infilling &Text only  &2.78 $\pm$ 0.48 & 13.89  $\pm$ 1.27 & 22.50 $\pm$ 1.44 & 2.48 $\pm$ 0.31 \\
    \midrule
    QuoteRetriever-T & Text only &\textbf{5.00} &  \textbf{17.50}  & \textbf{30.00} & \textbf{3.56 $\pm$ 0.22}\\   
    QuoteRetriever-TV & Text+Visual & \textbf{5.00}& 15.00 & 23.33 & 3.49 $\pm$ 0.26\\
    \bottomrule
  \end{tabular}
\end{table}

\subsection{Quote Retriever}\label{sec:retriver_eval}

In this experiment, we aim to compare the retrieval capability of our model against GPT-based infilling method. First, we report recalls for retrieving the correct interview segments given their preceding and succeeding narration of each teaser in our test set in \cref{retriver_gold}. On average, each video in our test set has 66 candidate interviews with a standard deviation of 37. Our proposed QuoteRetriever-T and QuoteRetriever-TV outperform infilling with GPT-4o and random selection. Please see \cref{sec:objective_eval_detail} for details about the objective evaluation on quote retrievers.

\begin{figure}
  \small
  \centering
  \begin{subfigure}[b]{0.32\textwidth}
    \includegraphics[width=\linewidth]{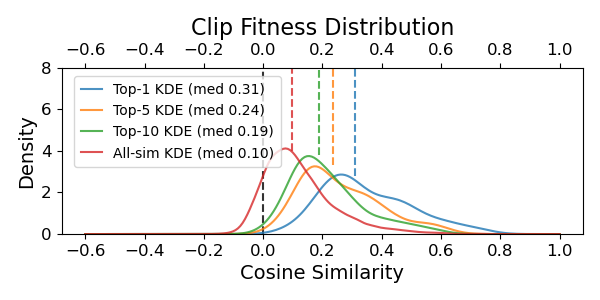}
    \vspace{-4ex}
    \caption{GPT-4o infilling}
    \label{fig:sub3}
  \end{subfigure}
  \hfill
  \begin{subfigure}[b]{0.32\textwidth}
    \includegraphics[width=\linewidth]{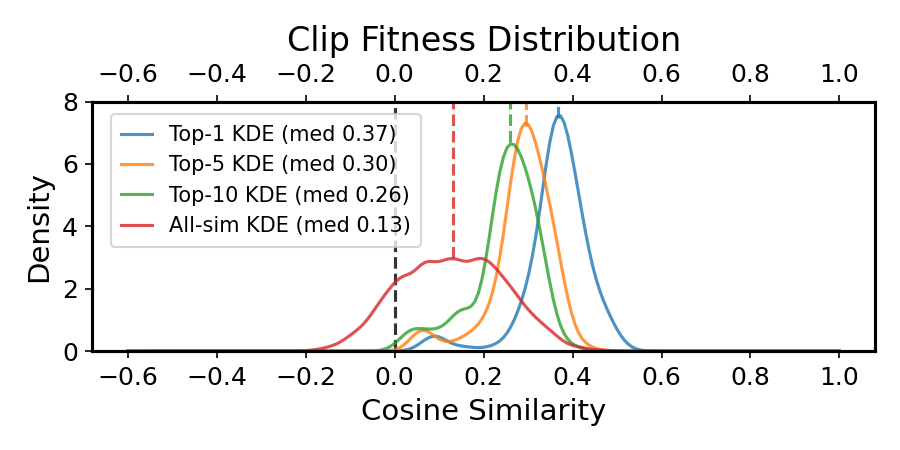}
    \vspace{-4ex}
    \caption{QuoteRetriever-T}
    \label{fig:sub1}
  \end{subfigure}
  \hfill
  \begin{subfigure}[b]{0.32\textwidth}
    \includegraphics[width=\linewidth]{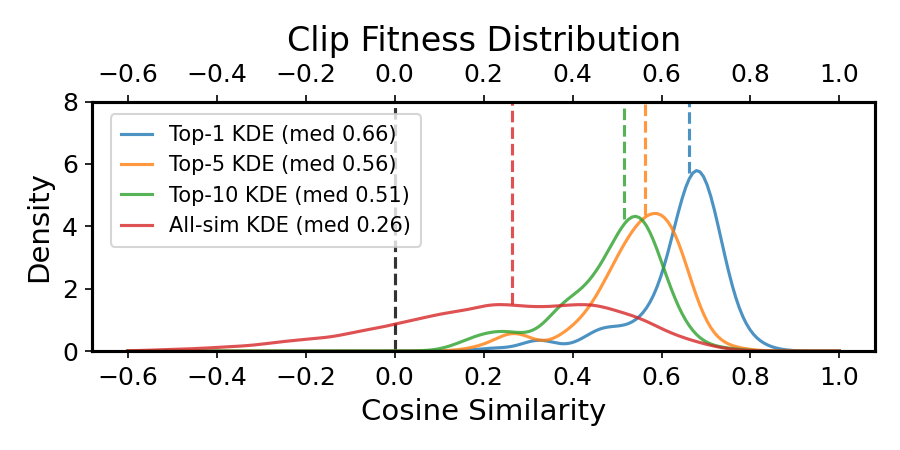}
    \vspace{-4ex}
    \caption{QuoteRetriever-TV}
    \label{fig:sub2}
  \end{subfigure}
  \caption{Comparison of infilling methods. The dotted lines indicate the median values.}
  \label{fig:infilling_distribution_plot}
\end{figure}

Second, to evaluate the discriminative capability of our retriever on interview segments, we plot in \cref{fig:infilling_distribution_plot} the top-1 retrieval similarity (``Top-1 KDE'') alongside the all similarity distribution (``All-sim KDE''), where `all' refers to the similarity between the embedding output by the fine-tuned BART\cite{lewis-etal-2020-bart} model and all interview segments. 
We observe a more prominent separation between the top-1 KDE and All-sim KDE for QuoteRetriever-TV than the GPT-4o infilling approach, which is beneficial for more accurate retrieval performance due to the increased contrast between probable candidates.

Third, we evaluate quote retriever performance by measuring how effectively retrieved segments can be inserted into generated narration scripts in a subjective study. We report the mean score and 95\% confidence interval in \cref{retriver_gold}. Both QuoteRetriever-T and QuoteRetriever-TV outperform the GPT-infilling baseline on interview effectiveness, indicating that pretrained GPT-4o cannot produce meaningful text to fulfill the \texttt{<QUOTE>} placeholders and support accurate retrieval.

\begin{table}
  \caption{Objective evaluation results for documentary teaser generation}
  \label{tab:teaser_obj}
  \centering
  \footnotesize
  \vspace{.5ex}
  \begin{tabular}{l@{~~~~}c@{~~~~}c@{~~~~}c@{~~~~}c@{~~~~}c@{~~~~}c@{~~~~}c@{~~~~}c@{~~~~}c}
    \toprule
    Model &\shortstack{Dur\\(sec)} &\shortstack{Interview\\ratio (\%)} &\shortstack{F1\\(\%)} &\shortstack{SCR\\(\%)} &\shortstack{REP\\(\%)} & VTGHLS & CLIPS-I & CLIPS-N \\
    \midrule
    Random extraction  &  101 & 56 $\pm$ 20 & 1.10 & 20.71 & 0.41 & 0.83 & 0.55 & 0.62\\
    ETS & 142  & 34 $\pm$ 16 & 1.92 & 13.65 & 4.49 & 1.06 & 0.64 & 0.60 \\
    A2Summ \cite{he2023alignattendmultimodalsummarization} & 73 & 42 $\pm$ 25 & 1.70 & 14.20 & 1.73 & 0.89 & 0.56 & 0.63 \\  
    TeaserGen \cite{xu2024teasergengeneratingteaserslong} &  155 & - & 1.64  &  \textbf{22.61} & 21.38 & 0.80 & - & 0.67\\
    GPT-4o-DQ & 151 & 42 $\pm$ 42 & 1.56 & 16.55 & 20.75 & 1.01 & 0.58 & 0.42 \\
    GPT-4o-SP-DQ & 619 & 61 $\pm$ 17 & \textbf{2.07} & 12.38 & 18.33 & 1.02  & 0.62 & 0.62 \\
    \midrule
    REGen-DQ &  95 & 37 $\pm$ 26 & 1.45 & 19.13 & 10.35 & 1.05 & 0.48 & 0.57 \\
    REGen-IDQ-T  &  77 & 35 $\pm$ 31 & 1.89 & 19.79 & 10.02 & 1.03  & \textbf{0.41} & \textbf{0.57} \\
    REGen-IDQ-TV & 81& 35 $\pm$ 31 & 1.90 & 19.86& \textbf{9.70} & 1.02 & 0.39 & 0.57 \\
    \midrule
    Ground truth &  76 & 54 $\pm$ 37 & 69.00\tabfn & 27.60 &  $>$ 7.86  & $<$0.98 & 0.43 & 0.57 \\
    \bottomrule
  \end{tabular}\\[.5ex]
  \scriptsize\raggedright\textsuperscript{*} Following \cite{xu2024teasergengeneratingteaserslong}, for each frame in the teaser, we retrieve the top 20 most similar frames from the main content using CLIP embeddings. We then apply pixel‐by‐pixel comparison to the 20 candidates ; however, this strict matching may fail to identify identical frames due to the low frame rate.
\end{table}

\subsection{Documentary Teaser Generation}

To evaluate the performance of the proposed method for the documentary teaser generation task, we compare our proposed method against several baselines models: random extraction, A2Summ \cite{he2023alignattendmultimodalsummarization}, TeaserGen \cite{xu2024teasergengeneratingteaserslong}, Extractive-then-Smoothing (ETS), GPT-4o-DQ, and GPT-4o-SP-DQ.
As reported in \cref{tab:objective_screenplay}, we can see that REGen-DQ achieves the highest ROUGE scores, indicating that REGen-DQ generates scripts closest to the ground-truth teasers. We also find that REGen-IDQ-TV achieves the closest G-Eval \cite{liu2023gevalnlgevaluationusing} score to that of the ground-truth,  
and that GPT-4o-SP-DQ achieves the highest F1 score. Notably, the CLIPScore for the interview scenes of the ground truth is lower than that of the narration scenes. This suggests a lower narration-visual correspondence for interview scenes, which is partly because interview scenes usually focus on the interviewees rather than having visuals to support the narration content. Our proposed REGen models also result in a smaller CLIPS-I value than CLIPS-N, which is consistent to the ground truth documentary teasers.
As can be seen from \cref{tab:teaser_subjective}, our subjective evaluation results show that the proposed REGen-IDQ-TV model achieves the highest scores in terms of coherence, alignment, and realness, outperforming our another proposed REGen-DQ model. Meanwhile, our proposed REGen-DQ method achieves the highest interview effectiveness score, but this does not reach significance difference against REGen-IDQ-TV.

For the extractive‐then‐smoothing (ETS) baseline, its higher VTGHLS score likely results from selecting interview chunks closest to the video title in sentence‐embedding space \cite{sentence-transformers-all-mpnet-base-v2}, but this value is higher than the ground‐truth VTGHLS.
Additionally, the ETS method has a much lower scene change rate than the ground truth teaser. In \cref{tab:objective_screenplay}, we observe that this method yields lower ROUGE scores than our proposed model, which indicates less overlap with the ground-truth teasers, while its higher G-Eval rating partly results from the enforced story-like prompt in \cref{sec:extractive_then_smoothing}. 
Second, A2Summ \cite{he2023alignattendmultimodalsummarization} yields the lowest G-Eval score in \cref{tab:objective_screenplay}, reflecting its low narrative cohesion, which is further verified by the lower coherence score in the subjective evaluation in \cref{tab:teaser_subjective}.
In \cref{tab:teaser_obj}, our model produces scene-change and repetition rates closer to those of the ground truth than A2Summ, which aligns with the higher perceived realness of our model in the subjective test.
Third, TeaserGen \cite{xu2024teasergengeneratingteaserslong} achieves the highest G-Eval score of 0.85 in \cref{tab:objective_screenplay}, indicating that it produces the most story-like and cohesive output; however, this value is higher than the ground truth value (0.62).
Also,
REGen-IDQ-TV achieves higher alignment than TeaserGen in the subjective test, suggesting that TeaserGen may select content that is less audio–visually aligned than naturally extracted interview segments. In addition, TeaserGen has the lowest VTGHLS score, suggesting that its retrieved frames are less likely to be considered as highlighted moments for its video title. 

\begin{table}
\centering
\footnotesize
\caption{Subjective evaluation results for documentary teaser generation}
\label{tab:teaser_subjective}
\vspace{.5ex}
\begin{tabular}{lccccccc}
\toprule
Model & Coherence$\uparrow$  & Alignment$\uparrow$ & Realness$\uparrow$ & Interview effectiveness$\uparrow$ \\ 
\midrule
A2Summ \cite{he2023alignattendmultimodalsummarization} & 2.72 $\pm$ 0.24 & 2.87 $\pm$ 0.26 & 2.67 $\pm$ 0.23 & 3.07 $\pm$ 0.24 \\
TeaserGen \cite{xu2024teasergengeneratingteaserslong} & 3.22 $\pm$ 0.23  & 2.92 $\pm$ 0.24 & 2.86 $\pm$ 0.23 & - \\
GPT-4o-SP-DQ & 3.08 $\pm$ 0.24 & 3.23 $\pm$ 0.25 & 2.81 $\pm$ 0.25 & 3.32 $\pm$ 0.25 \\
\midrule
REGen-DQ & 2.97 $\pm$ 0.27 & 3.03 $\pm$ 0.27 & 2.75 $\pm$ 0.30 & \textbf{3.33 $\pm$ 0.29}\\
REGen-IDQ-TV & \textbf{3.29 $\pm$ 0.24} & \textbf{3.30 $\pm$ 0.26} & \textbf{3.05 $\pm$ 0.25}& 3.25 $\pm$ 0.30 \\
\bottomrule
\end{tabular}
\end{table}



Finally, we compare our proposed model with GPT-based models, including GPT-4o-DQ and GPT-4o-SP-DQ.
We can see that the proposed REGen models produce scripts with higher ROUGE scores, indicating that the generated scripts are closer to the ground-truth teaser scripts. In the documentary teaser generation task, although GPT-4o-SP-DQ achieves the highest F1 score, it exhibits a much lower scene change rate and a significantly longer teaser length than the ground truth. Even though we cap teaser length at 500 tokens (approximately 2.5 minutes of speech assuming a normal speaking pace of 150 wpm), the generated teasers remain substantially longer than real ones. The lower scene change rate and higher repetitiveness suggest that GPT-4o-SP-DQ selects repetitive video clips, which can lead to a negative viewer experience. This aligns with the lower perceived realness of GPT-4o-SP-DQ compared to REGen-IDQ-TV in \cref{tab:teaser_subjective}. 
We also compare our model with GPT-4o-SP-TV, and the results can be found in \cref{sec:addition_teaser_gen}. 









\section{Limitations and Future Work}\label{sec:discussion_limitation}


While including video insertions may improve the factualness grounding of the output videos, the proposed method still has the risk of misplacing a quote in a wrong context. This may be alleviated by grounding the first-stage script generation model with information about all the quotable materials so that it can better generate a more cohesive narrative. To examine this hypothesis, 
we include in \cref{tab:quote_overlap} a baseline model (GPT-4o-DQ-NS) that supplies GPT-4o \cite{openai2024gpt4technicalreport} with the full transcript and all candidate quotable interview segments for script generation. However, this is technically challenging for LLaMA-based models due to the their limited context‐window \cite{touvron2023llamaopenefficientfoundation}, which prevents us from providing all the quote candidates as the context for script generation.
Finally, the proposed method relies on successful segmentation of the input video, which is possible in our case through speaker diarization that might not be applicable to other domains such as lecture recordings. For future work, we note that the proposed framework can be generalized to support quoting materials in any modality as long as we can find a proper fitness measure. We plan to investigate quoting audio and images towards a more capable video editing model.

\section{Conclusion}

We have proposed a novel retrieval-embedded generation framework that allows an LLM to include multimodal quotations in its outputs. We have examined the proposed method on the task of documentary teaser generation. We have shown through objective and subjective evaluations the effectiveness of our proposed REGen models on quoting short interview clips within a coherent narrative. The subjective evaluations show that our proposed hybrid approach outperforms several abstractive and extractive baseline models in terms of coherence, alignment, and realism.


\section{Acknowledgment}
This work is supported by the NVIDIA Academic Grant Program under the project titled "Teaser Generation for Long Documentaries and Educational Videos.
{
\small
\bibliographystyle{IEEEtran}
\bibliography{neurips_2025}
}



\newpage

\appendix

\section{Broader Impacts}\label{sec:broader_impacts}


We envision our proposed method to be applied to other fields such as education technology, natural language processing and information retrieval. For education technology, we believe our proposed model can be adapted to generate short review videos from lecture recordings to enhance learning experience. The proposed multimodal retrieval-embedded generation method can also be applied to current LLMs so that they can quote external sources embedded in their outputs. From the information retrieval perspective, our proposed method brings the power of LLMs to extractive methods where we can generate a coherent narrative to connect multiple extracted materials. We believe our proposed framework will contribute to improving knowledge accessibility by generating engaging short videos for long videos that may be more approachable to certain groups.

\section{Dataset Details}\label{sec:dataset_processing}

\paragraph{Generating Script with Speaker Annotation}\label{sec:script_gen}
In order to generate scripts with timestamp and speaker labeled, following \cite{mahon2024screenwriterautomaticscreenplaygeneration}, we use WhisperX \cite{whisperx} for speaker diarization, which generates a script with start, end timestamps, speaker IDs as well as the transcribed text. We assume that narration typically dominates and corresponds to the longest audio track; therefore, we label the speaker with the longest transcript as the narrator. An example of our processed data is available on our demo page.\cref{fn:demo}
We constructed 941 paired teaser–main documentary screenplays from DocumentaryNet \cite{xu2024teasergengeneratingteaserslong}, and we include an additional 157 teaser-only samples. 
We estimate that in our training and validation sets, 766 out of 1,098 teaser part contain inserted videos, with an average of 1.8 inserts per example. 

\paragraph{Validating Automatic Speech Recognition and Narrator Identification Results}

We recruited four people to evaluate the annotation quality of our dataset. We asked them to assess if the automatically detected narrator was correct, if the start and end times of an interview segment were accurate, and if the transcription achieved over 95\% accuracy. We provide an example of our annotation process on our demo page.\cref{fn:demo}
In the teaser part of our test set, we report the narrator prediction F1 score, audio-track start-time correctness, end-time correctness, and transcription accuracy. The narrator prediction F1 score is 71.6\%. Start-time correctness is 93.5\%. End-time correctness is 94.9\%. Transcription accuracy is 94.9\%. We present 128 interview segments with correct start times, end times, and transcriptions. In addition, 40 of 49 teasers in our test set include a second speaker, and on average each test video has 3.02 inserted clips.
In our main documentary part of our test set, we also report the narrator prediction F1 score, audio-track start-time correctness, end-time correctness, and transcription accuracy. 
The narrator prediction F1 score is 88.7\%. Start-time correctness is 90.8\%. End-time correctness is 91.7\%. Transcription accuracy is 96.3\%. We present 2,472 interview segments with human-validated start times, end times, and transcriptions.

\section{Implementation Details}\label{sec:implementation_detail}
All experiments are conducted on a single NVIDIA A100 GPU with a batch size of 64. We reserve 5 \% of the dataset (49 documentaries) for final testing and allocate 10 \% of the remaining samples for validation. The learning rate is set to $1\times10^{-5}$, and training terminates when either the generation or the retrieval validation loss does not decrease within 30 consecutive epochs. We use Adam optimizer for training. 

\subsection{Script Generation}\label{sec:appendix_script_gen}
In script generation stage, we remove any teaser outputs containing non-English tokens. Consequently, we fine-tune LLaMA \cite{touvron2023llamaopenefficientfoundation} on 839 paired examples.

\subsection{Quote Retriever}\label{sec:appendix_quote_retriever}
In the quote retriever stage of our pipeline, we construct 47,883 training samples to jointly fine-tune the all‐mpnet‐base‐v2 sentence embedder \cite{sentence-transformers-all-mpnet-base-v2} and the Facebook BART-base model \cite{lewis-etal-2020-bart} with a maximum input length of 256 tokens. This extended context window enables the retriever to capture long-range dependencies—such as narrative shifts and cross-segment entity references in long main documentaries—that often exceed 128 tokens. For the documentary teaser-generation task (\cref{sec:documentary_eval}) and the accompanying subjective evaluation (\cref{sec:subjective_survey}), we cap BART’s input window at 128 tokens. This choice is informed by our analysis in \cref{tab:objective_screenplay}, which shows that the typical teaser length is 121 tokens; thus, a 128-token limit tightly encompasses almost all real-world examples. Moreover, reducing the context window cuts inference time and memory usage by nearly half during large-scale A/B studies, while ensuring that all generated outputs are compared under identical input constraints.
In our GroupSampler module, if the number of distinct documentaries in a batch is less than the batch size, additional negatives are sampled from other documentaries to encourage fine‐grained discrimination.
When multiple interview segments occur between two consecutive narration chunks, we select the nearest preceding and succeeding narration scripts to construct each training sample.

\subsection{Documentary Teaser Generation}\label{sec:appendix_teasergen}
When constructing teasers with our proposed method, to prevent repetition, we maintain a sliding window over the last three selected clips and disallow any duplicate segment within that window.

\subsection{GroupSampler}\label{sec:appendix_groupsampler}
If the number of distinct documentaries in a batch is less than the batch size, additional negatives are sampled from other documentaries to encourage fine‐grained discrimination.

\section{Objective Evaluation Details}\label{sec:objective_eval_detail}

In \cref{tab:teaser_obj}, for each narration we may select multiple intervals (frames) to accompany it. We compute the CLIPScore between the narration script and each frame, then use the highest CLIPScore among them as the score for that narration. For each interview segment, we similarly consider the highest CLIPScore within its interval. We report CLIPS-I as the CLIPScore measuring audiovisual alignment of interview segments and CLIPS-N as the CLIPScore measuring audiovisual alignment of narration segments.

We observe that certain interview segments in the main documentary are post-processed (e.g., trimmed or reshoot) before being incorporated into teasers. We treat these edited segments as distinct items in our retriever stage. To ensure a fair comparison with fully extractive baselines, we remove those interviews that cannot be exactly found in main documentary when compute F1, Repetitivenss, CLIPScore and VTGHLS when evaluating teaser generation task in \cref{tab:teaser_obj}. There is no difference in SCR because one interview is usually considered as one scene. 

In \cref{retriver_gold}, we also notice that some of the teaser are fully narrations and those are removed for retriever stage evaluation. Moreover, we conduct the experiments with 3 random seeds and report the standard deviation for random selection and GPT-4o infilling. 

In \cref{tab:objective_screenplay}, we run G-Eval with three random seeds and report the mean score and standard deviation. 

\section{Script Generation Evaluation on Direct Quote from Long Contents}\label{sec:direct_quote_table}

\begin{table}[H]
  \caption{Comparison in the capability of direct quotation}
  \label{tab:quote_overlap}
  \footnotesize
  \centering
  \vspace{.5ex}
  \begin{tabular}{llccc}
    \toprule
    Model    & Quotation encoding &\shortstack[l]{Summarized\\input script} & \shortstack[l]{Semantic similarity with\\the nearest neighbor} & Overlap ratio  \\
    \midrule
    GPT-4o-DQ & Quotation marks & \yes & 0.52 & 0.07 \\
    GPT-4o-DQ-NS & Quotation marks & \no & 0.50 & 0.13 \\
    GPT-4o-SP-DQ  & Screenplay-like & \yes & 0.69 & 0.17 \\
    LLaMA-DQ & Quotation marks & \yes & - & 0.00 \\
    \midrule
    REGen-DQ & \texttt{<SOQ>} \& \texttt{<EOQ>} & \yes  & 0.45 & 0.07 \\ 
    \bottomrule
  \end{tabular}
\end{table}

In \cref{tab:quote_overlap}, we report the semantic similarity between segments predicted as quotations—that is, LLM outputs enclosed by quotation markers—and their nearest neighbors in our interview database, as well as the Overlap Ratio defined in \cref{sec:script_gen_experiment}. When we provide GPT-4o \cite{openai2024gpt4technicalreport} with the full documentary transcript, the overlap ratio is only 0.13. To accommodate extremely long inputs, we feed GPT-4o a summarized version of the main documentary; this further lowers the overlap to 0.07. The screenplay-like quotation encoding with GPT-4o raises the token overlap ratio to 0.17, but this remains inadequate. While vanilla LLaMA \cite{touvron2023llamaopenefficientfoundation} cannot produce meaningful quotation markers, fine-tuning it with \texttt{<SOQ>} and \texttt{<EOQ>} increases the overlap ratio from 0 to 0.07. However, we note that this overlap ratio is still lower than that of the GPT-4o-based model. We expect better performance if we scale up the dataset.

\section{Additional Teaser Generation Evaluation}\label{sec:addition_teaser_gen}
\begin{table}[H]
  \caption{Objective Evaluation for Teaser Generation Task}
  \label{tab:addition_teaser_gen}
  \centering
  \footnotesize
  \vspace{.5ex}
  \begin{tabular}{l@{~~~~}c@{~~~~}c@{~~~~}c@{~~~~}c@{~~~~}c@{~~~~}c@{~~~~}c@{~~~~}c@{~~~~}c}
    \toprule
    Model &\shortstack{Dur\\(sec)} &\shortstack{Interview\\Ratio (\%)} &\shortstack{F1\\(\%)} &\shortstack{SCR\\(\%)} &\shortstack{REP\\(\%)} & VTGHLS & CLIPS-I & CLIPS-N \\
    \midrule
    Random extraction  &  101 & 56 $\pm$ 20 & 1.10 & 20.71 & 0.41 & 0.83 & 0.55 & 0.62\\
    ETS & 142  & 34 $\pm$ 16 & 1.92 & 13.65 & 4.49 & 1.06 & 0.64 & 0.60 \\
    A2Summ \cite{he2023alignattendmultimodalsummarization} & 73 & 42 $\pm$ 25 & 1.70 & 14.20 & 1.73 & 0.89 & 0.56 & 0.63 \\  
    TeaserGen \cite{xu2024teasergengeneratingteaserslong} &  155 & - & 1.64  &  \textbf{22.61} & 21.38 & 0.80 & - & 0.67\\
    GPT-4o-DQ & 151 & 42 $\pm$ 42 & 1.56 & 16.55 & 20.75 & 1.01 & 0.58 & 0.42 \\
    GPT-4o-SP-DQ & 619 & 61 $\pm$ 17 & \textbf{2.07} & 12.38 & 18.33 & 1.02  & 0.62 & 0.62 \\
    GPT-4o-SP-TV & 673 & 64 $\pm$ 17& 1.61 & 11.29 & 41.46 & 1.02  & 0.64 & 0.62\\
    REGen-IDQ (random) &  82 & 32 $\pm$ 33 & 1.34 & 20.40 & \textbf{7.50} & 1.03& 0.41 & 0.57 \\
    \midrule
    REGen-DQ &  95 & 37 $\pm$ 26 & 1.45 & 19.13 & 10.35 & 1.05 & 0.48 & 0.57 \\
    REGen-IDQ-T  &  77 & 35 $\pm$ 31 & 1.89 & 19.79 & 10.02 & 1.03  & 0.41 & 0.57 \\
    REGen-IDQ-TV & 81& 35 $\pm$ 31 & 1.90 & 19.86& 9.70 & 1.02 & 0.39 & 0.57 \\
    \midrule
    Ground truth &  76 & 54 $\pm$ 37 & 69.00\tabfn & 27.60 &  7.86  & $<$0.98 & 0.43 & 0.57 \\
    \bottomrule
  \end{tabular}\\[.5ex]
  \scriptsize\raggedright\textsuperscript{*} Following \cite{xu2024teasergengeneratingteaserslong}, for each frame in the teaser, we retrieve the top 20 most similar frames from the main content using CLIP embeddings. We then apply pixel‐by‐pixel comparison to the 20 candidates; however, this strict matching may fail to identify identical frames due to the low frame rate.
\end{table}
When comparing GPT-4o-SP-DQ with GPT-4o-SP-TV, and GPT-4o-SP-TV with REGen-IDQ-TV, we observe a significant drop in teaser-generation performance for GPT-4o-SP-TV in \cref{tab:addition_teaser_gen}, indicating that the surrounding narration produced by our fine-tuned script aids the our retrieval stage.
\section{GPT-4o Prompts}\label{gpt_prompts}

\subsection{GPT-4o-DQ}\label{sec:gpt_4o_dq_prompt}
You are a helpful assistant. Generate an engaging introduction of the content with quotation based on the following input
\begin{verbatim}
Input: f"{Ten sentences chunk summary}"
\end{verbatim}
Output:

\subsection{LLaMA-DQ}\label{sec:llama_dq_prompt}
You are a helpful assistant. Generate an engaging introduction of the content with quotation based on the following input

\begin{verbatim}
Input: f"{Ten sentences chunk summary}"
\end{verbatim}
Output:

\subsection{GPT-4o-SP-DQ}\label{sec:gpt_4o_idq}
A. You are a helpful assistant. Generate an engaging introduction of the content with quotation based on the following input.

B. "Generate a concise version of the following screenplay segment by preserving its plain text format "
"where each line starts with either 'Narration:' or 'SpeakerID: [Speaker]:'. "
"Limit the summary to approximately 50 tokens. "

\subsection{LLaMA Finetuning Template for REGen-DQ}\label{sec:llama_finetune_regen_dq}
\begin{quote}
Instruction:
"You are a helpful assistant. Generate an engaging introduction of the content with quotation based on the following input."

\begin{verbatim}
Input: f"{Ten sentences chunk summary}"
\end{verbatim}

Output: 
\begin{align}
    &&\dots, x_k, \texttt{<SOQ>}, y_1, \dots, y_n, \texttt{<EOQ>}, x_{k+1}, \dots
\end{align}
\end{quote}

\subsection{LLaMA Finetuning Template for REGen-IDQ}\label{sec:llama_finetune_regen_idq}
\begin{quote}
Instruction:
"You are a helpful assistant. Generate an engaging introduction of the content with quotation based on the following input."

\begin{verbatim}
Input: f"{Ten sentences chunk summary}"
\end{verbatim}

Output: 
\begin{align}
&&\dots, x_k, \texttt{<QUOTE>}, x_{k+1}, \dots
\end{align}
\end{quote}

\subsection{G-Eval}\label{sec:g_eval_prompt}
\begin{quote}
``Assess how naturally the text flows in a story-like manner, evaluating grammatical correctness, syntactic variety, and seamless transitions that enhance narrative coherence.''
\end{quote}

\subsection{Extraction-then-Smoothing}\label{sec:extractive_then_smoothing}
System:
You are a creative storyteller and writing coach.

User:
Process the following input. It contains two interview chunks, each prefixed with “Interview:”. Treat each “Interview: …” segment as a single, indivisible unit and do not split or merge them.

\begin{verbatim}
Input: f"{Selected Interviews}"
\end{verbatim}

Your task:
1. Generate a concise, engaging story that preserves the exact wording of each interview chunk.
2. Clearly indicate where in the narrative each interview chunk is inserted by using [Interview].
Now, please craft the story with these guidelines.
"""


\section{Survey Questions}\label{survey_questions}

\subsection{Demographic Questions}

\begin{itemize}
    \item How often do you watch documentary?
    \item Do you have any video editing experience?
    \item On a scale of 1–5, where 1 = Beginner and 5 = Native/Bilingual, how would you rate your English proficiency?
\end{itemize}

\subsection{Interview Insertion Evaluation}

Please indicate your level of agreement with the following statement: \textit{``The inserted interview integrates with the narration, effectively supports the surrounding claim, and maintains a natural flow.''}

\subsection{Documentary Teaser Generation}

\begin{itemize}
    \item \textbf{Coherence}: To what extent do you feel that the sample maintains coherence and a smooth flow, ensuring that each segment transitions logically and the overall experience feels seamless?
    \item \textbf{Alignment}: To what extent do you feel that the narration and video match and work well together, making the overall presentation clear and easy to follow?
    \item \textbf{Realness}: How well do you feel this sample meets your expectations as a teaser for a documentary in general?
    \item 
    \textbf{Interview Effectiveness}: Please indicate your level of agreement with the following statement: \textit{``The inserted interview integrates with the narration, effectively supports the surrounding claim, and maintains a natural flow.''}
\end{itemize}

\subsection{Generalizability Evaluation Prompts}

\begin{itemize}
    \item \textbf{Lecture Videos}: Which sample would you prefer as a teaser for lecture videos?
    \item \textbf{News Videos}: Which sample would you prefer as a teaser for news videos?
\end{itemize}

\subsection{Generalizability Evaluation Results}\label{sec:general_eval}

To assess the generalizability of our framework, we randomly select 10 lecture videos from the Multimodal Lecture Presentations Dataset \cite{lee2022multimodallecturepresentationsdataset}, covering subjects such as psychology, machine learning, dentistry, and biology, and we also include full-episode news broadcasts from NBC News. We then conduct an A/B study comparing our method against TeaserGen \cite{xu2024teasergengeneratingteaserslong}, asking participants which teaser they would prefer for each lecture video or news broadcast.  We perform a side-by-side evaluation against TeaserGen \cite{xu2024teasergengeneratingteaserslong} on both lecture and news videos. In the lecture domain, participants prefer TeaserGen 62\% of the time versus 38\% for our model, indicating a clear preference for abstractive summaries there. For news videos, the split is 55\% in favor of TeaserGen and 45 \% for our approach—a smaller gap that does not reach significance.

\begin{table}[H]
\centering
\footnotesize
\caption{Subjective evaluation results of Generated Teaser}
\label{tab:narr_subjective}
\vspace{.5ex}
\begin{tabular}{lcc}
\toprule
Dataset & Model  &  Preference (\%) \\
\midrule
News & REGen-IDQ-TV & 45 \\
News &TeaserGen & 55 \\
\cmidrule(lr){1-3}
Lecture videos & REGen-IDQ-TV & 38 \\
Lecture videos &TeaserGen & 62 \\
\bottomrule
\end{tabular}
\end{table}

\section{Ablation Study}\label{sec:ablation_study}


\subsection{Effects of Max Context Window Length}\label{sec:ablation_context_window}

We compare different context window length in \cref{tab:retriver_len}. We find no significant difference when we set different context window as the teasers are usually short. 

\begin{table}[H]
  \caption{Effects of maximum context window length}
  \label{tab:retriver_len}
  \footnotesize
  \centering
  \vspace{.5ex}
  \begin{tabular}{lllll}
    \toprule
    Model   & Max context tokens & Recall@1 (\%)    & Recall@5 (\%) & Recall@10 (\%)   \\
    \midrule
    GPT-4o infilling & 128 &  2.78 $\pm$ 0.48 & 13.89  $\pm$ 1.27 & 22.50 $\pm$ 1.44 \\
    GPT-4o infilling & 256 & 3.33 $\pm $0.83 & 13.33 $\pm$ 0.83 & 22.78 $\pm$ 1.27 \\
    QuoteRetriever-T &128 &  5.00 & 15.00 & 23.33 \\
   QuoteRetriever-T & 256& 4.17 & 16.67 & 22.50 \\
   QuoteRetriever-TV &128 &\textbf{5.00} &  \textbf{17.50}  & \textbf{30.00}\\   
  QuoteRetriever-TV &256 & \textbf{5.00} &  \textbf{17.50}  & \textbf{27.50}\\
    \bottomrule
  \end{tabular}
\end{table}

\subsection{Effects of Alpha in Balancing Loss} \label{sec:ablation_alpha_loss}

We compare in \cref{alpha_effect} the effects of the weights of the generation loss and retrieval loss in the loss function \cref{eq:loss_func}. We find that when $\alpha = 1$, meaning that generation loss and retrieval loss are equally weighted, our model achieves its highest performance. In the following table, we set include 30\% interviews in a batch as hard negative samples. Here we set max length of tokens for each sample being 256. 

\begin{table}[H]
  \caption{Effects of $\alpha$ in the loss function \cref{eq:loss_func}}
  \label{alpha_effect}
  \small
  \centering
  \vspace{.5ex}
  \begin{tabular}{llll}
    \toprule
    $\alpha$     & Recall@1 (\%)     & Recall@5 (\%) & Recall@10 (\%) \\
    \midrule
    0 &  0.00  & 6.67  &  14.17 \\
    0.5    & 2.50  &15.83  &  29.17 \\
    1   &2.50   &  20.00 & 32.50 \\
    2     & 5.00 & 18.33     & 30.83 \\
    \bottomrule
  \end{tabular}
\end{table}

\subsection{Effects of Group Sampler} \label{sec:ablation_groupsampler}

To examine whether treating interviews from the same documentary as hard negatives improves retriever training, we conduct two experiments. In the first, we treat all interviews within the same documentary as hard negative samples. In the second, we treat only 30 \% of those interviews as hard negatives and omit the group sampler. We find that the 30 \% group-sampler configuration yields the highest recall@5 and recall@10, while omitting the group sampler achieves the highest recall@1.

\begin{table}[H]
  \caption{Effect of negative sampler construction and loss function at $\alpha = 1$}
  \label{tab:negativesampler}
  \centering
    \small
    \vspace{.5ex}
  \begin{tabular}{llcccc}
    \toprule
    Model &  Loss  & GroupSampler   & Recall@1 (\%)     & Recall@5 (\%) & Recall@10 (\%) \\
    \midrule
    Model-Visual& Contrastive & \yes &  5.00 &  17.50  & 27.50 \\
    Model-Visual& L2 & \yes & 1.67  &  6.67& 13.33 \\
    Model-Visual & Contrastive & 30\% & 2.50 & 20.00 & 32.50 \\
    Model-Visual & Contrastive & \no & 7.50 & 17.50 & 31.67 \\
    \bottomrule
  \end{tabular}
\end{table}

\subsection{Effect of Loss Function} \label{sec:ablation_loss_func}

We also use $L_{2}$ loss to find the closest embedding during the retrieval stage (see \cref{tab:negativesampler}). We find that leveraging contrastive loss increases our model’s performance, yielding higher recall. This is likely because contrastive loss can better differentiate embeddings that are close in the embedding space. Here we set max length of tokens for each sample being 256. 

\subsection{Effect of Position of Retrieval token} \label{sec:ablation_retrival_token_position}
As shown in \cref{tab:retriver_token}, appending the special token \texttt{<SUM>} to the end of the decoder output before retrieval improves recall. Here we set max length of tokens for each sample being 256. 

\begin{table}[H]
  \caption{Effect of the position of the \texttt{<SUM>} token}
  \label{tab:retriver_token}
  \small
  \centering
  \vspace{.5ex}
  \begin{tabular}{llll}
    \toprule
    Position     & Recall@1 (\%)     & Recall@5 (\%) & Recall@10 (\%) \\
    \midrule
    Start &  1.67 & 10.00   &  25.83  \\
    End    & 2.50  & 20.00  &  32.50 \\
    \bottomrule
  \end{tabular}
\end{table}

\section{Comparison within REGen System}\label{sec:regen_comparesion}
We compare the models in REGen system with different variants. 
We find that REGen-DQ achieves the highest ROUGE score and the most realistic quote distribution, as indicated by the quote coverage rate and quotation density index, both closest to the ground truth. The G-Eval scores for REGen-IDQ-T and REGen-IDQ-TV are closer to the ground truth than REGen-DQ, indicating that they produce more coherent, story-like scripts under automatic LLM evaluation.
In \cref{tab:teaser_subjective}, our subjective evaluation further indicates that our proposed models, REGen-IDQ-T and REGen-IDQ-TV, receive higher ratings for interview effectiveness compared with REGen-IDQ (random), indicating the effectiveness of our proposed retriever. 
In \cref{tab:teaser_obj}, we present the effects of different retriever methods in the documentary teaser-generation task, we find that REGen-IDQ-TV achieves the highest F1 score among models in REGen system. 
In \cref{tab:teaser_subjective} our subjective evaluation of teaser generation shows that teasers generated by REGen-DQ yield higher interview-effectiveness scores than those by REGen-IDQ-TV, indicating that fine-tuning to enable direct quotations can increase the coherence and supportiveness of inserted interviews.

\end{document}